# Investigating Deep-Learning NLP for Automating the Extraction of Oncology Efficacy Endpoints from Scientific Literature

Gendrin-Brokmann[1], Aline, Harrison, Eden*[1], Noveras, Julianne*[1], Souliotis, Leonidas*[1], Vince, Harris*[1], Smit, Ines[2], Costa, Francisco[2], Milward, David[2], Dimitrievska, Sashka [1], Metcalfe, Paul[1], Louvet, Emilie[1]

*equal contribution

1. AstraZeneca, Cambridge and Gaithersburg

2. Linguamatics, an IQVIA business, based in Cambridge, UK; www.linguamatics.com

## Abstract

### Objective

Benchmarking drug efficacy is a critical step in clinical trial design and planning. The challenge is that much of the data on efficacy endpoints is stored in scientific papers in free text form, so extraction of such data is currently a largely manual task. Our objective is to automate this task as much as possible.

### Methods

In this study we have developed and optimised a framework to extract efficacy endpoints from text in scientific papers, using a machine learning approach.

### Results

Our machine learning model predicts 25 classes associated with efficacy endpoints and leads to high F1 scores (harmonic mean of precision and recall) of 96.4% on the test set, and 93.9% and 93.7% on two case studies.

### Conclusion

These methods were evaluated against – and showed strong agreement with – subject matter experts and show significant promise in the future of automating the extraction of clinical endpoints from free text.

### Significance

Clinical information extraction from text data is currently a laborious manual task which scales poorly and is prone to human error. Demonstrating the ability to extract efficacy endpoints automatically shows great promise for accelerating clinical trial design moving forwards.

## Graphical Abstract

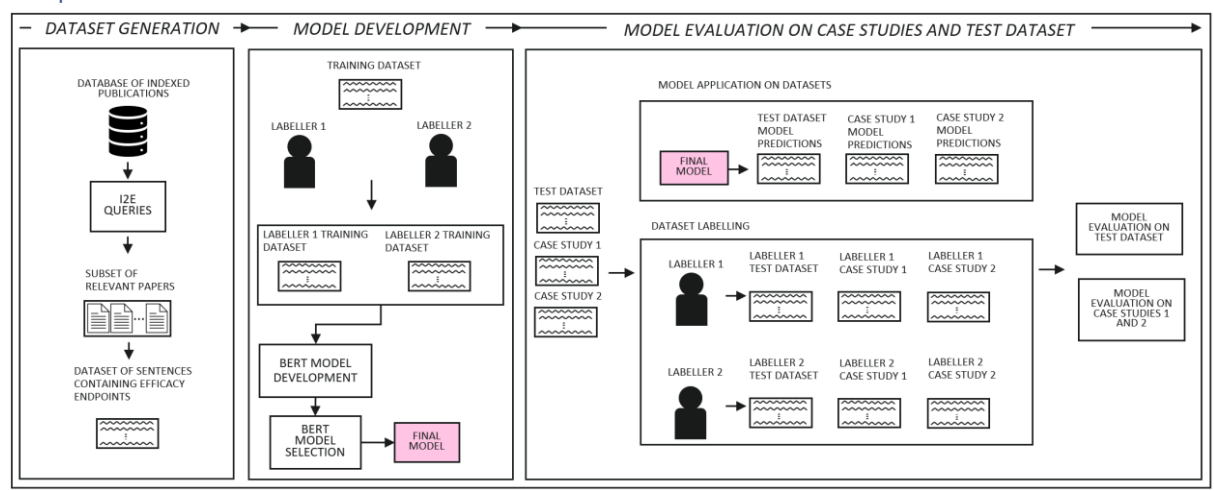

*Figure 1: Graphical abstract. This high-level overview summarises the three main steps involved in predicting efficacy endpoints in scientific papers.*

## Introduction

Benchmarking efficacy of a new drug against drugs in development and already available on the market is essential for trial planning and design, providing the necessary data to facilitate improved understanding of treatment effectiveness and disease cohorts. The challenge, however, is that most of the data appears in clinical trial reports or scientific journal articles as unstructured text[1] which is often only accessible by reading the text to find the relevant information.

### Overview of Information Retrieval and Information Extraction

Quantitative literature reviews can be divided into two steps: information retrieval and information extraction. Information retrieval involves retrieving the relevant subset of articles related to a topic of interest. It must be performed with a high degree of specificity and is becoming increasingly difficult as volumes of published literature increase year on year[2]–[4]. Retrieval of scientific papers[5] can be made more efficient by designing highly refined search queries. Less advanced search tools are also available which retrieve documents in databases containing scientific journals and clinical texts such as TrialTrove, PubMed and Google Scholar which facilitate relatively basic inclusion and exclusion criteria. However, these often do not allow for more refined queries capable of retrieving documents containing specific sequences of text or searching for words based on their linguistic class, associations, or synonyms. Information extraction[6] involves identifying the relevant data – such as efficacy endpoints – from within each article; in this case study, it requires a very low false negative rate. Achieving this manually is not efficient or scalable[7] and introduces the possibility of human error[8], [9].

### Automating Information Extraction

The repetitive nature of information extraction means it is ripe for automation through the use of natural language processing (NLP). NLP appears frequently in the clinical domain and there are many examples where it has been applied to free-text data from scientific literature. NLP has huge potential to assist with drug development[10], clinical trial design[11], and understanding of best evidence in primary healthcare[12]. NLP can be used for information extraction, which involves indexing and pre-processing of the text dataset, extraction of information, and usually interpretation of such information[13]. Information extraction can be broadly divided into two categories: rule-based and machine learning. Rule based approaches involve designing a set of linguistic patterns which extract data depending on whether a sequence of characters matches that particular pattern, whereas machine learning approaches use statistical approaches to extract relevant data.

Clinical information extraction is often seen in electronic medical records[13]–[17] as well as in scientific literature[6], [18]–[20]. It has also been applied in several other domains, including materials science[21] and chemical information extraction[22]. Rule-based information extraction has seen popularity applied to clinical texts [8], [23], [24]. There are several software packages which have been developed and used in healthcare and clinical contexts that enable rule-based extraction such as i2e [25], which has been used for information extraction in clinical contexts[26] as well as for scientific abstracts[27].

Deep-learning based transformer models such as BioBERT[28], PubMedBERT[29], ClinicalBERT[30], BlueBERT[31], SciBERT[32] have demonstrated success on information extraction tasks in clinical corpora for a range of tasks including Named Entity Recognition (NER). BERT has been used for clinical information extraction for several tasks [33].

## Contribution

We have developed a machine learning approach to extract efficacy endpoints from scientific journal articles, as detailed in Figure 1. Using a rule-based query language within i2e, we developed datasets tailored to our specific problem from the tens of millions of sentences within MEDLINE, and pre-highlighted them with entities of interest. After manual correction and annotation, the fully labelled dataset was used to train and optimise a set of machine learning models based on BERT, before selecting the best model. Finally, we evaluated the performance of the best model on two case studies which consisted of literature searches requiring efficacy endpoint extraction from a subset of selected scientific journal articles.

## Materials and Methods

### Dataset Generation and Annotation

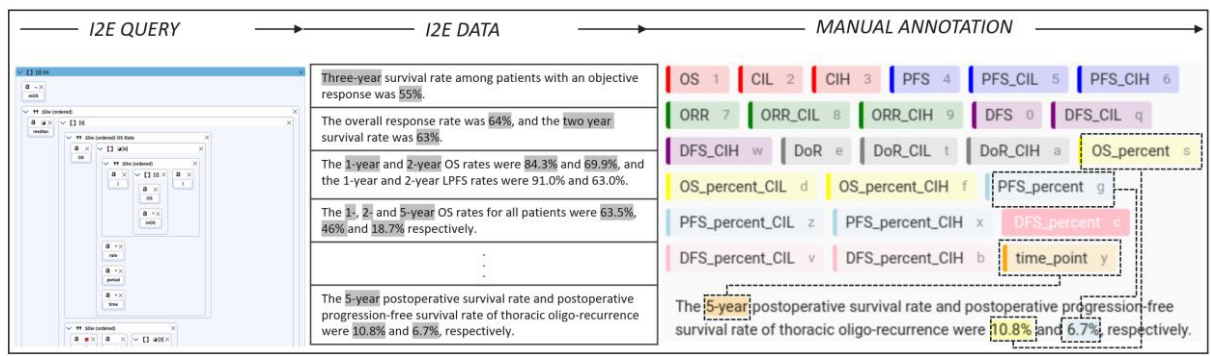

*Figure 2: Training and test dataset generation and annotation process. Numerical values are highlighted by i2e to assist the labeller when annotating entities within each sentence. Annotations are corrected upon manual review. CIL=Confidence Interval Low, CIH=Confidence Interval High. "Percent" refers to entities recorded as percentages.*

| Entity Name (Preferred Term) | Abbreviation |
| --- | --- |
| **Overall Survival** | OS |
| **Overall Response Rate** | ORR |
| **Progression Free Survival** | PFS |
| **Disease Free Survival** | DFS |
| **Duration of Response** | DoR |
| **Time Point** | - |

*Table 1: Entities to be predicted. A total of 25 variations of these endpoints are included in the prediction task, since entities are treated as different endpoints if they are expressed as a duration or a percentage. Confidence interval bounds are also considered as separate endpoints. See in Supplementary Material for more explanations.*

To train, optimise and evaluate the machine learning models, two datasets were generated using an index of MEDLINE in i2e[25] – a training dataset and a test dataset containing 5392 and 983 individual sentences respectively. The MEDLINE dataset is a bibliography of articles in the life science and biomedical domain. I2e is effective for creating machine learning datasets, enabling sufficient variability in semantic phrasing styles, sentence lengths and frequency of occurrence of each of the endpoints to be captured. For traceability, each sentence was stored along with the PubMed ID of the paper that it came from. The training dataset was used for model training and hyperparameter optimisation. The test dataset was used to evaluate the predictive performance of the model. Two case study datasets were also generated using i2e, each consisting of 70 and 37 abstracts / full texts from scientific journal articles respectively. The case studies were used to evaluate the model performance on a dataset representative of what would be used in a real literature review.

Efficacy endpoints in peer-reviewed papers can be expressed in different ways with different units, such as percentages and durations, which sometimes have associated confidence intervals. It was important to include this distinction and so we added additional classes for the different ways each endpoint could be expressed. For example, OS_percent corresponds to overall survival when it is expressed as a percentage, and OS, corresponds to overall survival when it is expressed as a duration. High and low confidence intervals result in two additional entities per endpoint. For example, OS_CIL corresponds to Confidence Interval Low for Overall Survival, while OS_CIH corresponds to Confidence Interval High for Overall Survival. Table 2 and Table 3 show that our classes are highly imbalanced. This reflects their frequency of occurrence in peer-reviewed literature. Generally, confidence intervals appear a lot less frequently than the corresponding endpoint. We have almost 3000 words corresponding to DFS entities, while we have only 1000 words for the corresponding lower confidence interval (DFS_CIL) and approximately the same number of words for the higher confidence interval (DFS_CIH). Also, there are more words for some entities than for others. There are almost 7000 DFS entities expressed as a percentage (DFS_percent), while there are only approximately 2000 DoR entities.

For the training dataset, an ensemble of queries was built that would each focus on one endpoint. Most sentences in the training dataset contain at least one occurrence of either of the entities in Table 1. Sometimes, when the endpoint appears more rarely in the literature, several queries were built with slightly different constraints until the number of entities per endpoint reached a sufficiently high value to lead to high cross validated F1 scores. Some blocks in some of these queries were built by a linguistic expert thus ensuring that the different linguistic constructions which appear in the peer-reviewed literature around the mention of an endpoint value are present in our dataset. An example of such a block is shown in the left-hand side of Figure 2. The query will consider sentences with either mOS, or median OS plus optionally OS or mOS between parentheses, with the words rate, period or time as alternative possibilities. Additional cases are also covered that are not shown in Figure 2, because of space limitation. The training dataset contained several negative training samples to avoid confusion over semantically similar terms. Negative training samples consisted of unlabelled sentences containing entities that could be mistaken for one of the endpoints in Table 1 and were deliberately included in order for the models to learn not to predict them as false positives. Potential false positive examples were age, which could be confused with survival, length of stay (LOS) which could be confused with overall survival (OS), the phrase "median duration" could be confused with duration of response (DoR), standard deviation values could be confused with confidence intervals, values reported as percentages could be confused as ORR.

The test dataset was built using the simplest possible query that would cover the endpoints from Table 1. This query requests that one of the endpoints would be present in the sentence, along with a corresponding numerical entity, consisting of a duration and/or percentage. Any PMIDs (PubMed Identifiers) from the training dataset were excluded from the test dataset.

Case study datasets correspond to real life literature searches and consist of a set of PubMed abstracts or full texts. Only a few sentences in the abstracts and full texts contain values that we are interested in, so we selected just these sentences for the case study dataset. To do so, we used the combination of a high-recall i2e query that will extract all sentences with values or percentages and one mention of the endpoints in Table 1, as well as the fine-tuned BERT model.

| Total number of sentences | Number of words in Training Dataset entities 5392 sentences | Number of words in Test Dataset entities 983 sentences |
|---|---|---|
| DFS | 2907 | 56 |
| DFS_CIH | 1282 | 0 |
| DFS_CIL | 1176 | 0 |
| DFS_percent | 6847 | 1654 |
| DFS_percent_CIH | 1980 | 73 |
| DFS_percent_CIL | 1770 | 67 |
| DoR | 1931 | 23 |
| DoR_CIH | 677 | 6 |
| DoR_CIL | 737 | 6 |
| ORR | 3320 | 509 |
| ORR_CIH | 541 | 104 |
| ORR_CIL | 503 | 95 |
| OS | 5051 | 1092 |
| OS_CIH | 1418 | 198 |
| OS_CIL | 1372 | 180 |
| OS_percent | 5805 | 5489 |
| OS_percent_CIH | 2063 | 379 |
| OS_percent_CIL | 1880 | 343 |
| PFS | 5139 | 536 |
| PFS_CIH | 1332 | 143 |
| PFS_CIL | 1256 | 121 |
| PFS_percent | 1758 | 353 |
| PFS_percent_CIH | 959 | 24 |
| PFS_percent_CIL | 981 | 16 |
| time_point | 5925 | 2591 |

*Table 2: This shows the total number of sentences in the training and test datasets as well as the total number of endpoints occurring across all sentences averaged across datasets from the two different labellers. Endpoints can be expressed either as a duration or as a percentage and they are considered as two different endpoints. Confidence intervals are also treated as separate endpoints.*

| | Number in Case Study 1 | Number in Case Study 2 |
|---|---|---|
| Total abstracts/full texts: | 70 | 37 |
| DFS | 0 | 0 |
| DFS_CIH | 0 | 0 |
| DFS_CIL | 0 | 0 |
| DFS_percent | 0 | 49 |
| DFS_percent_CIH | 0 | 8 |
| DFS_percent_CIL | 0 | 4 |
| DoR | 22 | 4 |
| DoR_CIH | 12 | 0 |
| DoR_CIL | 10 | 0 |
| ORR | 248 | 219 |
| ORR_CIH | 66 | 6 |
| ORR_CIL | 62 | 8 |
| OS | 367 | 152 |
| OS_CIH | 123 | 15 |
| OS_CIL | 105 | 15 |
| OS_percent | 59 | 5 |
| OS_percent_CIH | 10 | 0 |
| OS_percent_CIL | 10 | 0 |
| PFS | 363 | 158 |
| PFS_CIH | 136 | 10 |
| PFS_CIL | 117 | 8 |
| PFS_percent | 44 | 0 |
| PFS_percent_CIH | 11 | 0 |
| PFS_percent_CIL | 11 | 0 |
| time_point | 62 | 26 |

*Table 3: This shows the total number of abstracts or full texts in the two case study datasets as well as the total number of words in endpoints occurring across all abstracts averaged across datasets from the two different labellers.*

## Data Description

There are numerous examples which illustrate the types of sentence constructions present in the data, which highlight the difficulty of the machine learning task as well as the labelling task.

One notable feature of this dataset is a very high frequency of what we called "respectively" constructions. These are patterns in the text, often co-occurring with the word "respectively", where two lists of items are described separately, and the items in those lists are to be paired up according to their position within their list. Among these "respectively" constructions, there are several sub-patterns relating to which items are listed. There are many examples, such as

- A list of time points related to a list of values: "**3-, 5-** and **10-year** survival rates were **45%, 40%** and **35%**, respectively".
- A list of values related to a list of reporting or treatment groups: "Estimated **1-year** PFS rates were **78.2%** (95% CI **70.2-84.2**) and **83.0%** (95% CI **75.0-88.6**) for PF-05280586 and rituximab-EU, respectively."

The patterns furthermore differ in how they combine with the remaining items to be extracted. For instance, for the class of sub-patterns involving a list of efficacy endpoints and a list of values:

- Several values of a same endpoint are present in the sentence: "The **5-year** overall survival (OS) and disease-free survival rates were as follows: normal group, **82.5** and **76.8%;** emphysema group, **80.0** and **74.9%;** fibrosis group, **46.9** and **50%;** and CPFE group, **36.9** and **27.9%**, respectively (p < 0.01)."
- The same sentence can also contain another endpoint described via a different construction: "The objective response rate was **97.8%**; the median progression-free survival and OS were **11.0** and **27.0** months, respectively."
- There can be a second list of values, associated with a different time point, but the list of efficacy endpoints is not repeated: "The **1-year** LRFS, distant metastasis-free survival, disease-free survival, and overall survival rates were **78.2%, 78%, 69.8%**, and **90.2%**, respectively; the **3-year** rates were **50.6%, 41.2%, 31.2%**, and **66.3%**, respectively."

Another notable class of patterns involves comparisons in the text. These are similar to the "respectively" constructions, but instead of lists (sequences delimited by "," and "and" and of arbitrary length) we find pairs of items separated by "than" or "versus". They typically involve a comparison of reporting groups and a comparison of values, but the way they are combined with the remaining items to be extracted can again differ, e.g.:

- A comparison of reporting groups followed by a list whose elements are each formed by one outcome measure type (and associated time point) and a comparison of values:
  "Both disease-free survival (DFS) and overall survival (OS) were significantly worse in the CD8-Low/FoxP3-High group than the other groups (**5-year** DFS: **66.3%** vs. **90.5%**; P = 0.0007, **5-year** OS: **90.9%** vs. **97.0%**; P = 0.0077)."
- The outcome measure type precedes the two comparisons:
  "overall survival in all randomly assigned patients was significantly longer in the experimental group than in the control group (median **14·1 months** [95% CI **13·2-16·2**] vs **10·7 months** [**9·5-12·4**]"

Of course, many of the instances involve simpler sentence structures, e.g.:

- "**7-year** PFS of FL patients on RB therapy was **70%** (95% CI **75-99**)"

## Labelling

Each sentence was manually labelled for every occurrence of the entities in Table 1 using label-studio[34]. This was performed independently by two clinical information subject matter experts. Any sentence containing disagreements between labellers was reviewed, which helped identify any mislabelling instances so that they could be reconciled. Usually the two labellers agree, but on some occasions they do not. Hence, the reconciliation process is used in such cases to help labellers reach agreement. Once reconciled, the agreement between the two labellers was 99.9% for the training dataset, 99.9% for the test dataset, 99.9% for case study 1, and 99.7% for case study B. Two examples where the wording could be misleading are seen here:

- A partial response to AG with a mean duration of 9 months (range: 4-26 months ) was achieved in 24 patients (33% ), 10 patients (14% ) had stable disease, and 32 patients (44% ) were progressing during AG therapy."
- "According to the intent-to-treat analysis, 14/58 objective responses (**24.1%** ) and 24/58 (41.3% ) stabilizations of disease were observed, with a median duration of **4 months** (range, 2-22 + months ) and 5 months (range, 1-13 months ), respectively."

In both sentences the duration is not clearly stated as *Duration of Response*, however in the second sentence it is a duration of response, since it contains the phrase "objective response". There were also several additional examples where labelling was ambiguous, as in this example of two lists (2 reporting groups and 2 time points) that create a 4-fold combination to be matched against 4 values:

- "In the propensity score matched lobectomy and segmentectomy groups (87 patients per group), the **5-year** and **10-year** OS and PFS rates were **85%** versus **84%** and **66%** versus **63%,** respectively."

Note that depending on how it is interpreted you could match the second value ("84%") to the first endpoint (OS) or to the second endpoint (PFS). We can assume the first interpretation is the intended one on the basis that the first two values are close to each other.

## Data availability

All sentences used for designing queries and model training, testing, and evaluation on case studies were taken from the MEDLINE abstract database. The MEDLINE dataset is freely accessible through PubMed. For the two case studies presented, the data were downloaded using the PMC File Transfer Protocol. Only the papers with licence for commercial use were retrieved.

## Machine Learning Methods for Efficacy Endpoint Detection

We investigated four widely used state of the art transformer-based models including BERT[35], distilBERT[36], BioBERT[28] and PubMedBERT[29]. BERT is a bidirectional encoder based on the transformer[35], [37] architecture which was trained on a dataset containing billions of words (the full Wikipedia and BooksCorpus content). It can be fine-tuned for a variety of NLP-based tasks such as classification and named entity recognition (NER). BioBERT uses the same architecture as BERT and was fine-tuned on biomedical corpora from PubMed and PMC[28]. As such, it was optimised for texts in the clinical domain. PubMedBERT is very similar to BioBERT, but it was trained from scratch on PubMed and PMC without any initialisation rather than being fine-tuned on these datasets[29] . DistilBERT is a compressed version of BERT which retains 95% of the original performance but is significantly faster to fine-tune[38].

The machine learning task took the form of a multiclass (25-class) NER problem. The model training procedure started with tokenising the data using Hugging Face's AutoTokenizer module, which splits sentences into individual tokens and then assigns a label to each token, where a token is a single word or part of a word. The four different transformer models based on BERT[35] which are listed above were initialised on pretrained weights and fine-tuned on the two separate training datasets from each labeller. The code for the fine-tuning process was adapted from the Hugging-Face library. Models were evaluated using 5-fold cross validation. The seed was fixed to allow for reproducibility.

## Results

## Machine Learning Model Performance on Training and Test Datasets

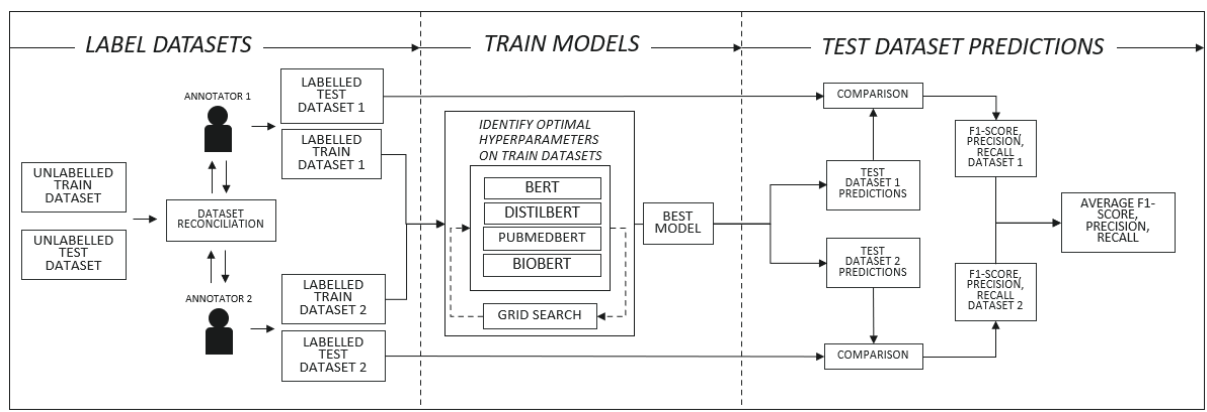

*Figure 3: Model training and evaluation procedure. The test and training datasets are annotated by each labeller. The labelled training datasets are then used to fine-tune the four models (BERT, DistilBERT, PubMedBERT and BioBERT) with the model achieving the highest average F1-score, precision, recall being selected for predictions on the test datasets.*

The model selection procedure (detailed in Figure 3) involved fine-tuning each of the four models on all 18 combinations of hyperparameters in Table 4 using grid search as described in Devlin *et al.*[35]. Models were scored based on the average precision, recall and F1-score achieved for each of the predicted entities. Precision is a measure of how often a predicted entity exactly matches the labelled data, recall is a measure of how many of the labelled entities were identified, and F1-score is the harmonic mean of precision and recall. The optimal set of hyperparameters for all models are shown in bold in Table 5. It happens in this case that all models were optimized with the same hyperparameters. The selected model was BioBERT, which had the highest F1-score, precision and recall although it shows only marginally better scores than PubMedBERT and takes twice as long to train as DistilBERT. This difference in computational time is not a discriminative factor for the case study presented here but we note for the future that a gain of a factor of two in computational time can be achieved here if needed, with only a small drop in F1-score (3.1%). All the models perform well, and the overall scores are generally high, achieving an average F1-score of 95.3% across all of the fine-tuned BERT-based models.

| Hyperparameter | Possible Values |
| --- | --- |

| | | |
|---|---|---|
| Epochs | 2, 3, **4** | |
| Learning rate | 2e-5, 3e-5, **5e-5** | |
| Batch size | **16**, 32 | |

*Table 4: Model hyperparameters explored. Optimal hyperparameters for all models are highlighted in bold.*

| Model | F1-score | Precision | Recall | Application Time (s) |
|---|---|---|---|---|
| **BERT** | 95.1 | 94.9 | 95.2 | 513 |
| **distilBERT** | 93.3 | 93.3 | 93.2 | **275** |
| **BioBERT** | **96.4** | 96.2 | **96.6** | 482 |
| **PubMedBERT** | 96.4 | **96.3** | 96.5 | 508 |

*Table 5: Performance metrics of BERT variant models on training dataset with optimised hyperparameters. Best scores for each column are highlighted in bold.*

A breakdown of the 5-fold cross-validated scores achieved on the training dataset on each endpoint is shown in Table 6, for the selected BioBERT-based model. In a NER task, F1 scores are of particular importance, as NER problems are generally associated with high class imbalance. Table 6 shows very high cross validated F1 scores, and are above 95% for most endpoints, except for PFS_percent (PFS when it is expressed as a percentage in the text), PFS_percent_CIL (Confidence Interval Low for PFS when it is expressed as a percentage) and PFS_percent_CIH (Confidence Interval High for PFS when it is expressed as a percentage), where scores are closer to 90%. Such high values of F1 scores necessarily correspond to high precision and recall values (because F1 is the harmonic mean of precision and recall). Precision and recall consistently display similarly high values.

| Endpoint | F1-score | Precision | Recall |
|---|---|---|---|
| **DFS** | 96.9 | 96.5 | 97.3 |
| **DFS_CIH** | 96.7 | 96.9 | 96.5 |
| **DFS_CIL** | 97.0 | 97.3 | 96.7 |
| **DFS_percent** | 96.5 | 96.1 | 97.0 |
| **DFS_percent_CIH** | 95.7 | 95.8 | 95.7 |
| **DFS_percent_CIL** | 95.6 | 95.5 | 95.7 |
| **DoR** | 97.2 | 97.0 | 97.4 |
| **DoR_CIH** | 95.3 | 93.3 | 97.4 |
| **DoR_CIL** | 97.8 | 97.1 | 98.4 |
| **ORR** | 97.8 | 98.3 | 97.3 |
| **ORR_CIH** | 95.7 | 96.6 | 95.2 |
| **ORR_CIL** | 96.9 | 98.0 | 96.1 |
| **OS** | 97.0 | 96.8 | 97.2 |
| **OS_CIH** | 96.5 | 96.2 | 96.7 |
| **OS_CIL** | 96.4 | 96.0 | 96.8 |
| **OS_percent** | 96.4 | 96.3 | 96.6 |
| **OS_percent_CIH** | 96.0 | 96.0 | 96.1 |
| **OS_percent_CIL** | 95.2 | 95.8 | 94.6 |
| **PFS** | 97.7 | 97.3 | 98.2 |
| **PFS_CIH** | 97.2 | 96.8 | 97.6 |
| **PFS_CIL** | 97.0 | 96.8 | 97.2 |
| **PFS_percent** | 90.8 | 89.0 | 92.8 |
| **PFS_percent_CIH** | 94.8 | 94.1 | 95.5 |
| **PFS_percent_CIL** | 89.5 | 93.4 | 86.0 |
| **time_point** | 96.8 | 96.4 | 97.3 |
| **overall** | 96.4 | 96.2 | 96.6 |

*Table 6: Cross-validated performance metrics of BioBERT model on training dataset. Overall scores are the average of each column weighted by the abundance of each entity in that row.*

Table 7 displays the scores for the test dataset. An important point to note is that there is a significant class imbalance in the datasets, however this is reflective of the abundance of different endpoints in the literature. For example, there are no instances of Disease-Free Survival confidence intervals (DFS_CIL and DFS_CIH) when they are expressed as a duration. Other classes that only count a few instances are DoR and confidence intervals for PFS expressed as a percentage (PFS_CIL and PFS_CIH). Most F1 scores are above 90% and all F1 scores are above 85%.

| Endpoint | F1-score | Precision | Recall |
|---|---|---|---|
| **DFS** | 96.6 | 93.3 | 100 |
| **DFS_percent** | 95.1 | 95.3 | 95.0 |
| **DFS_percent_CIH** | 87.2 | 85.5 | 89.0 |
| **DFS_percent_CIL** | 87.4 | 86.8 | 88.1 |
| **DoR** | 88.0 | 81.5 | 95.7 |
| **DoR_CIH** | 92.3 | 85.7 | 100 |
| **DoR_CIL** | 100 | 100 | 100 |
| **ORR** | 84.7 | 80.6 | 89.2 |
| **ORR_CIH** | 97.6 | 97.1 | 98.1 |
| **ORR_CIL** | 96.9 | 95.9 | 97.9 |
| **OS** | 96.7 | 97.4 | 96.1 |
| **OS_CIH** | 97.2 | 96.5 | 98.0 |
| **OS_CIL** | 98.3 | 98.3 | 98.3 |
| **OS_percent** | 97.6 | 97.7 | 97.4 |
| **OS_percent_CIH** | 97.4 | 97.6 | 97.2 |
| **OS_percent_CIL** | 97.7 | 97.4 | 98.0 |
| **PFS** | 97.7 | 98.1 | 97.2 |
| **PFS_CIH** | 98.3 | 97.3 | 99.3 |
| **PFS_CIL** | 98.4 | 97.6 | 99.2 |
| **PFS_percent** | 96.4 | 94.8 | 98.0 |
| **PFS_percent_CIH** | 90.9 | 100 | 83.3 |
| **PFS_percent_CIL** | 85.7 | 100 | 75.0 |
| **time_point** | 96.7 | 95.5 | 98.1 |
| **overall** | 96.4 | 96.0 | 96.8 |

*Table 7: Performance metrics using the BioBERT model, averaged across the test datasets labelled by labeller 1 and labeller 2. Overall scores are the average of each column weighted by the abundance of each entity in that row.*

## Performance on Case Studies

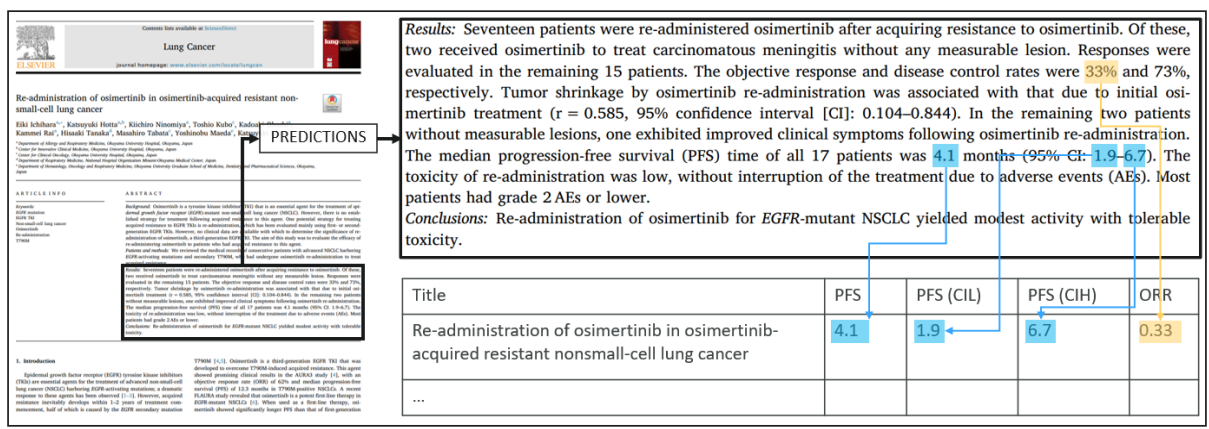

*Figure 4: Schematic of model and query predictions on case study data. Efficacy endpoint predictions are made using the BERT models and/or queries.*

The model was then evaluated on the two case study datasets (Figure 4). The case studies involved extracting the efficacy endpoints for two separate literature searches. The dataset for each case study consisted of the abstract or full text of each of the papers selected as part of the literature review and i2e was used to select only sentences which contain a duration and/or percentage. All endpoints occurring in the resulting set of sentences were manually and independently annotated by the two labellers, the labels were then reconciled. Following this, the fine-tuned BioBERT model was used to predict the entities.

|  | F1-Score | Precision | Recall |
|---|---|---|---|
| **DoR** | 95.7 | 91.7 | 100 |
| **DoR_CIH** | 85.7 | 100 | 75.0 |
| **DoR_CIL** | 90.9 | 83.3 | 100 |
| **ORR** | 93.0 | 90.2 | 96.0 |
| **ORR_CIH** | 89.6 | 94.9 | 84.8 |
| **ORR_CIL** | 89.1 | 93.0 | 85.5 |
| **OS** | 97.0 | 94.1 | 100 |
| **OS_CIH** | 97.1 | 97.5 | 96.7 |
| **OS_CIL** | 97.2 | 96.3 | 98.1 |
| **OS_percent** | 95.1 | 99.1 | 91.5 |
| **OS_percent_CIH** | 100 | 100 | 100 |
| **OS_percent_CIL** | 100 | 100 | 100 |
| **PFS** | 94.0 | 91.6 | 96.4 |
| **PFS_CIH** | 88.1 | 100 | 78.7 |
| **PFS_CIL** | 87.7 | 94.1 | 82.1 |
| **PFS_percent** | 100 | 100 | 100 |
| **PFS_percent_CIH** | 62.5 | 100 | 45.5 |
| **PFS_percent_CIL** | 62.5 | 100 | 45.5 |
| **time_point** | 100 | 100 | 100 |
| **overall** | 93.9 | 94.2 | 93.6 |

*Table 8: Performance metrics using the BioBERT model, averaged across the case study 1 dataset labelled by labeller 1 and labeller 2. Overall scores are the average of each column weighted by the abundance of each entity in that row.*

Table 8 and Table 9Table 8 report the results of the fine-tuned BioBERT model on case study 1 and case study 2 respectively. Scores are generally very high, mostly above 85%, demonstrating that the method provides informative results for the user. When scores are below 85%, namely for the upper and lower boundaries of PFS values when they are expressed as percentages (PFS_percent_CIH and PFS_percent_CIL), the number of considered entities is low, namely 11 entities in each of PFS_percent_CIH and PFS_percent_CIL for case study 1. A similar effect is observed in case study 2, where there are only 5 instances of OS_percent, resulting in a low F1-score of 66.7%.

|  | F1-Score | Precision | Recall |
|---|---|---|---|
| **DFS_percent** | 97.9 | 100 | 96.0 |
| **DFS_percent_CIH** | 85.7 | 100 | 75.0 |
| **DFS_percent_CIL** | 71.7 | 57.1 | 100 |
| **DoR** | 100 | 100 | 100 |
| **ORR** | 91.0 | 92.4 | 89.5 |
| **ORR_CIH** | 100 | 100 | 100 |
| **ORR_CIL** | 100 | 100 | 100 |
| **OS** | 98.4 | 96.8 | 100 |
| **OS_CIH** | 100 | 100 | 100 |
| OS_CIL | 100 | 100 | 100 |
| OS_percent | 66.7 | 57.1 | 80.0 |
| PFS | 92.4 | 97.2 | 88.0 |
| PFS_CIH | 100 | 100 | 100 |
| PFS_CIL | 100 | 100 | 100 |
| time_point | 85.1 | 95.2 | 76.9 |
| **overall** | **93.7** | **95.2** | **92.3** |

*Table 9: Performance metrics using the BioBERT model, averaged across the case study 2 dataset labelled by labeller 1 and labeller 2. Overall scores are the average of each column weighted by the abundance of each entity in that row.*

## Discussion

Automated information extraction in scientific literature reviews is of crucial importance for clinical benchmarking and clinical trial design. To date, this task is performed manually, incurring significant time penalty, and introducing the potential of human error. We have developed a fine-tuned BERT based method for extracting efficacy endpoints as well as associated timepoints from scientific papers. The best model achieved excellent performance with 5-fold cross-validated F1-scores of 96.4%. The F1 score on the test set was 96.4% as well. We applied our efficacy endpoint extraction techniques on two separate case-studies, where we aimed to identify efficacy endpoints from within a set of abstracts or full-text papers retrieved for real business cases. The average F1-scores for case study datasets 1 and 2 were also very high, 93.9% and 93.7%, respectively.

We identified two related research articles which aimed to extract and identify endpoints in clinical texts. Mutinda *et al.*[39] developed a BERT-based model to identify clinical outcomes – more specifically PICO (Participants, Intervention, Control and Outcomes) information from PubMed abstracts. This was treated as a NER task and they identified the PICO elements achieving an F1-score greater than 80% for most entities. But the entities were not matched by type of endpoints, all endpoint values were placed in the same NER category; separating the endpoints by type would require an additional relation extraction step which was not performed as part of the research. Kehl *et al.* [40] used deep-learning NLP to extract oncologic outcomes from radiology reports and treated the task as 8 binary classification problems, e.g. cancer progression vs no progression. They assessed patient reports at regular intervals, and as an example, the date of the first report classified as 'progression' would mark the end of progression-free survival. Our work searched for a larger variety of endpoints and identified the type of endpoint each value referred to in each case.

From a usability perspective, the Hugging-Face library was straightforward to implement for a data scientist but would not be straightforward to implement for a non-technical user and there is a need to put the model in production. Identifying the relevant cohort or treatment group corresponding to an endpoint was not considered here, as the problem is addressed separately [41].

Machine-learning models, especially deep-learning models are sometimes thought to lack explainability [42]–[46]. In this work, outputs of the machine learning model are reviewed by a subject matter expert (SME), thus eliminating possible biases. Indeed, the outputs of the model are used to highlight endpoint values in peer-reviewed papers to facilitate the work of the SME. The SME reviews all the highlighted endpoint values, which often correspond to several patient populations. The SME also double checks that no endpoint has been missed by the model. The SME will then select only the highlighted endpoint values corresponding to the patient population of interest for a particular study. It is important to note that before applying the model for information extraction, an information retrieval step has been made where peer-reviewed papers have been carefully selected, as described in the introduction. Additionally the selected endpoints are carefully checked by SMEs and any discrepancies are scrutinized, in the same way as when the information is extracted manually.

Further improvements could be made to the machine learning approach moving forward. It would be necessary to add additional endpoints, particularly with the changing landscape of clinical development driving the use of new and different endpoints. Adding an endpoint to the existing model is cumbersome, as the entire training dataset needs to be relabelled for this additional endpoint. However, there is a workaround which consists in cascading several models that each focus on a limited ensemble of endpoints and this would be straightforward to implement for the machine learning approach. The number of endpoints cannot be too small for each model as it is proven that multitask learning improves results for NER tasks[47].

Finally, this work shows that BERT-based models can be applied effectively to problems where the correct classification depends on complex patterns in the text, as in this case many of the relevant examples come from sentences exhibiting "respectively" or comparison constructions, or even combinations of the two.

## Conclusion

A fine-tuned BERT model was proposed as a way of automatically extracting efficacy endpoint information from biomedical literature. This method was evaluated against – and showed strong agreement with – subject matter experts and shows significant promise in the future of automating the extraction of clinical endpoints from free text.

## Acknowledgements

We thank AstraZeneca and Linguamatics (IQVIA NLP) for funding this project.